\documentclass[pdflatex,sn-mathphys]{sn-jnl}

\jyear{2021}%

\theoremstyle{thmstyleone}%
\usepackage{amsmath}
\usepackage{amsfonts}

\usepackage{multirow}
\usepackage{booktabs}
\usepackage{longtable}

\usepackage{subcaption}

\usepackage[numbers]{natbib}

\usepackage{pifont}
\newcommand{\xmark}{\ding{53}}

\theoremstyle{thmstyletwo}%

\theoremstyle{thmstylethree}%

\begin{document}

\title[]{End-to-End Optimized Image Compression with the Frequency-Oriented Transform}








\author[1]{\fnm{Yuefeng} \sur{Zhang}}\email{yuefeng.zhang@pku.edu.cn}

\author[2]{\fnm{Kai} \sur{Lin}}\email{kai.lin@pku.edu.cn}



\affil[1]{
\orgname{Beijing Institute of Computer Technology and Application}, \orgaddress{\street{51th Yongding Road, Haidian District}, \postcode{100039}, \state{Beijing}, \country{China}}}

\affil[2]{\orgdiv{School of Computer Science}, \orgname{Peking University}, \orgaddress{\street{5th Yiheyuan Road, Haidian District}, \postcode{100871}, \state{Beijing}, \country{China}}}



\abstract{

Image compression constitutes a significant challenge amidst the era of information explosion. 
Recent studies employing deep learning methods have demonstrated the superior performance of learning-based image compression methods over traditional codecs. However, an inherent challenge associated with these methods lies in their lack of interpretability. 
Following an analysis of the varying degrees of compression degradation across different frequency bands, we propose the end-to-end optimized image compression model facilitated by the frequency-oriented transform.
The proposed end-to-end image compression model consists of four components: spatial sampling, frequency-oriented transform, entropy estimation, and frequency-aware fusion. The frequency-oriented transform separates the original image signal into distinct frequency bands, aligning with the human-interpretable concept. Leveraging the non-overlapping hypothesis, the model enables scalable coding through the selective transmission of arbitrary frequency components.
Extensive experiments are conducted to demonstrate that our model outperforms all traditional codecs including next-generation standard H.266/VVC on MS-SSIM metric.
Moreover, visual analysis tasks (i.e., object detection and semantic segmentation) are conducted to verify the proposed compression method could preserve semantic fidelity besides signal-level precision.
}




\maketitle

\section{Introduction}

With the explosion of image content on the internet, image compression technology shows its importance in information transmission and storage. Generation after generation, conventional image compression standards were proposed from JPEG \cite{wallace1992jpeg}, JPEG 2000 \cite{rabbani2002jpeg2000}, High Efficient Coding (HEVC)/H.265 \cite{sullivan2012overview} to the latest Versatile Video Coding (VVC)/H.266 \cite{bross2021overview} which was finalized in July 2020. The compression field focuses on rate-distortion (R-D) optimization to use as few bits as possible to represent the original image while keeping distortion below an acceptable range, which has led to stable but incremental progress.

With the fast development of neural networks, learning-based image compression methods \cite{rippel2017real,balle2018variational,minnen2018joint,cheng2020image,hu2021learning} tend to surpass traditional codecs, abandoning the traditional coding architecture (i.e., hybrid codec). Due to high commercial potential, industry companies developed their own learning-based image/video solutions, e.g. Google \cite{balle2016end}, NVIDIA\footnote{https://developer.nvidia.com/maxine} and Alibaba\footnote{https://segmentfault.com/a/1190000040968923/en}.
Data-driven codecs are proposed to learn the image distribution from big data and eliminate data probability estimation errors for various image contents. End-to-end optimized codec has shown its coding efficiency potential but lacks intuitive interpretability. In this regard, it is necessary to design a interpretable learning-based image compression code.

\begin{figure}[t]
  \centering
  \includegraphics[width=0.6\linewidth]{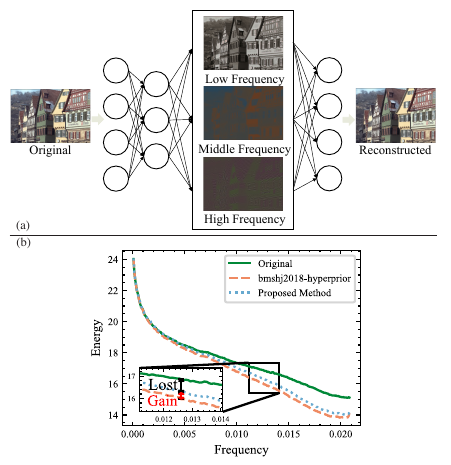}
  \caption{(a) Conceptual illustration of our proposed end-to-end optimized image compression model with the frequency-oriented transform that original image signal is tranformed into several frequency splits to further eliminate redundancy. (b) Power spectral density distribution chart through the Fourier transform which compares the degradation caused by different compression methods. }
  
\label{fig:frequency_display}
\end{figure}

Reviewing traditional lossy image codes, all of them follow three main steps: transform, quantization, and entropy coding. The transform step aims at removing the spatial redundancy by decorrelating the coefficient and energy compaction. Frequency-oriented decomposition has shown great success in traditional codecs (e.g., discrete cosine transform (DCT) in JPEG \cite{wallace1992jpeg} and discrete wavelet transform (DWT) in JPEG2000 \cite{rabbani2002jpeg2000}), which has also caused much research attention in the learning-based computer vision methods \cite{li2020learning,akbari2020generalized}. The visual signal can be split into different frequency bands through frequency-oriented decomposition that low spatial frequencies correspond to features such as global shape, while high spatial frequencies connect to aspects like sharp edges and fine details \cite{imagehandbook}. It has been proved that in the human visual system (HVS) eyes react differently according to the contents composed of different spatial frequencies \cite{imagehandbook,antonini1992image}. Moreover, recent experiments on the visual system of pigeons proved that higher spatial frequency is more important to recognize \cite{murphy2015pigeons}, showing a close connection between visual perception and image spatial frequency distribution. Thus, frequency-oriented decomposition has backup from engineer application and bio-science.

To examine image compression from the aspect of spectral energy distribution, we analyze the image power spectral density by the Fourier transform as shown in Fig.~\ref{fig:frequency_display}(b). It can be found out image compression causes different image degradation degrees on different frequency bands that image suffers much degradation on the high-frequency band compared with on the low-frequency one. This result is consistent with the design of classic codecs to reduce visual redundancy that HSV is not sensitive to high-frequency components and hence cuts off much higher frequency information than the lower one. 

These findings verify that bridging frequency analysis with image compression has the HSV theory basis. Here comes the question: how could we take the help of the properties of HVS and give learning-based image compression models interpretability?


\begin{figure*}
    \centering 
    \begin{subfigure}[b]{1\columnwidth}
        \centering
        \includegraphics[width=0.3\textwidth]{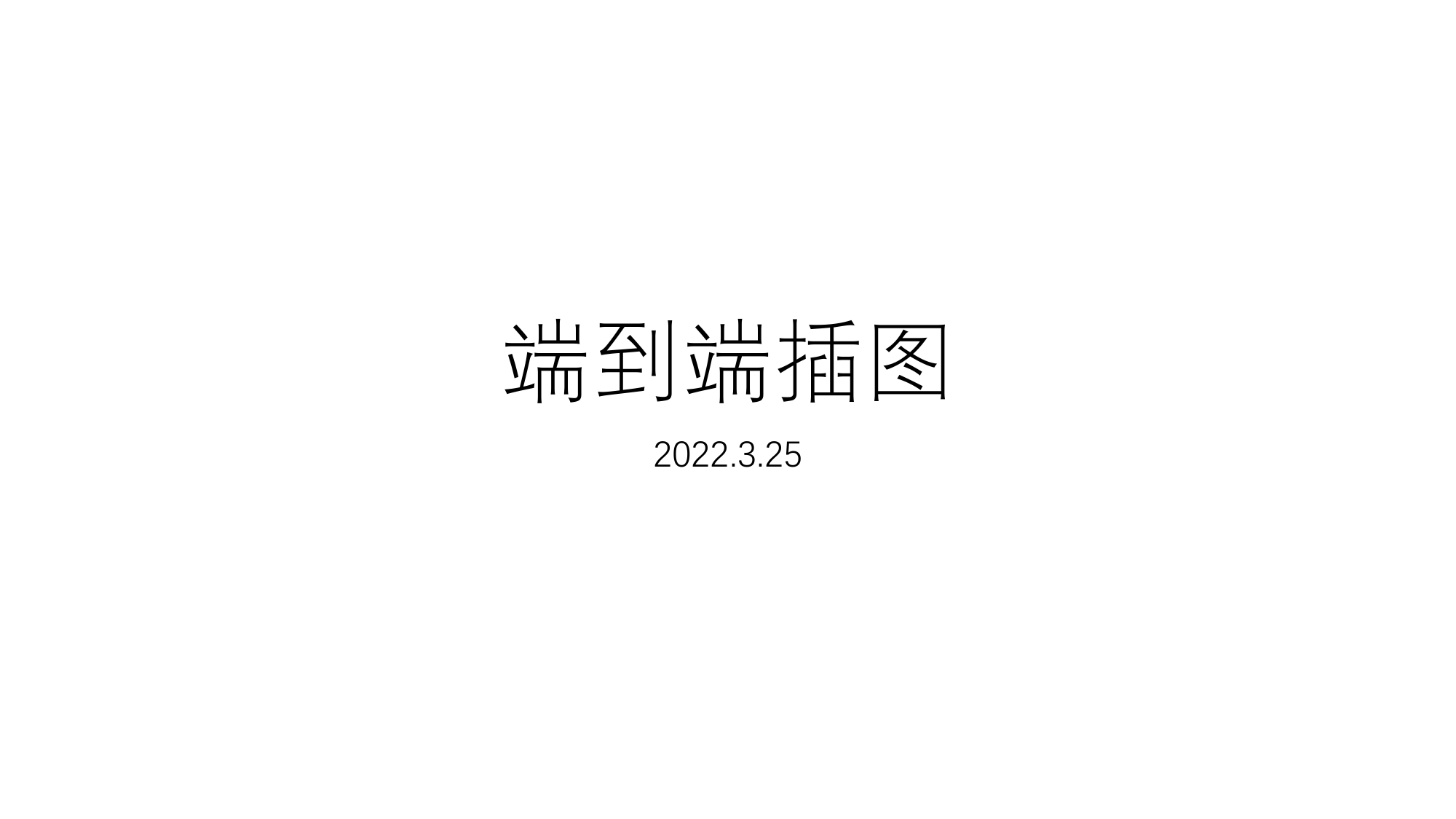}
        \caption{Baseline transform architecture which is commonly used in learning-based image compression models \cite{balle2016end,balle2018variational}. }
        \label{fig:arch1}
    \end{subfigure}
    \hfill
    \begin{subfigure}[b]{1\columnwidth}
        \centering
        \includegraphics[width=0.5\textwidth]{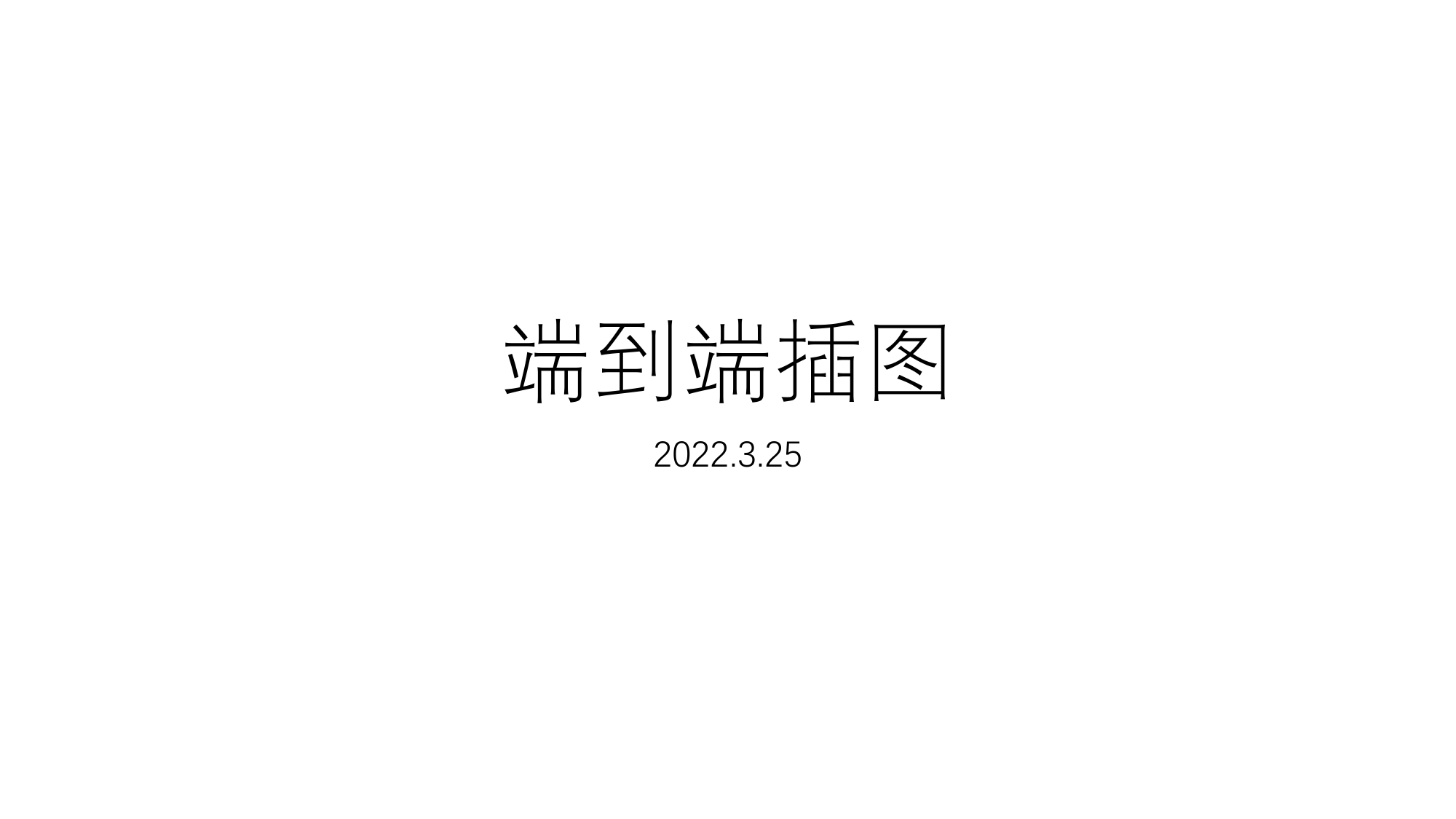}
        \caption{Separate transform and same entropy estimation module in \cite{rippel2017real}.}
        \label{fig:arch2}
    \end{subfigure}

    \begin{subfigure}[b]{1\columnwidth}
        \centering
        \includegraphics[width=0.5\textwidth]{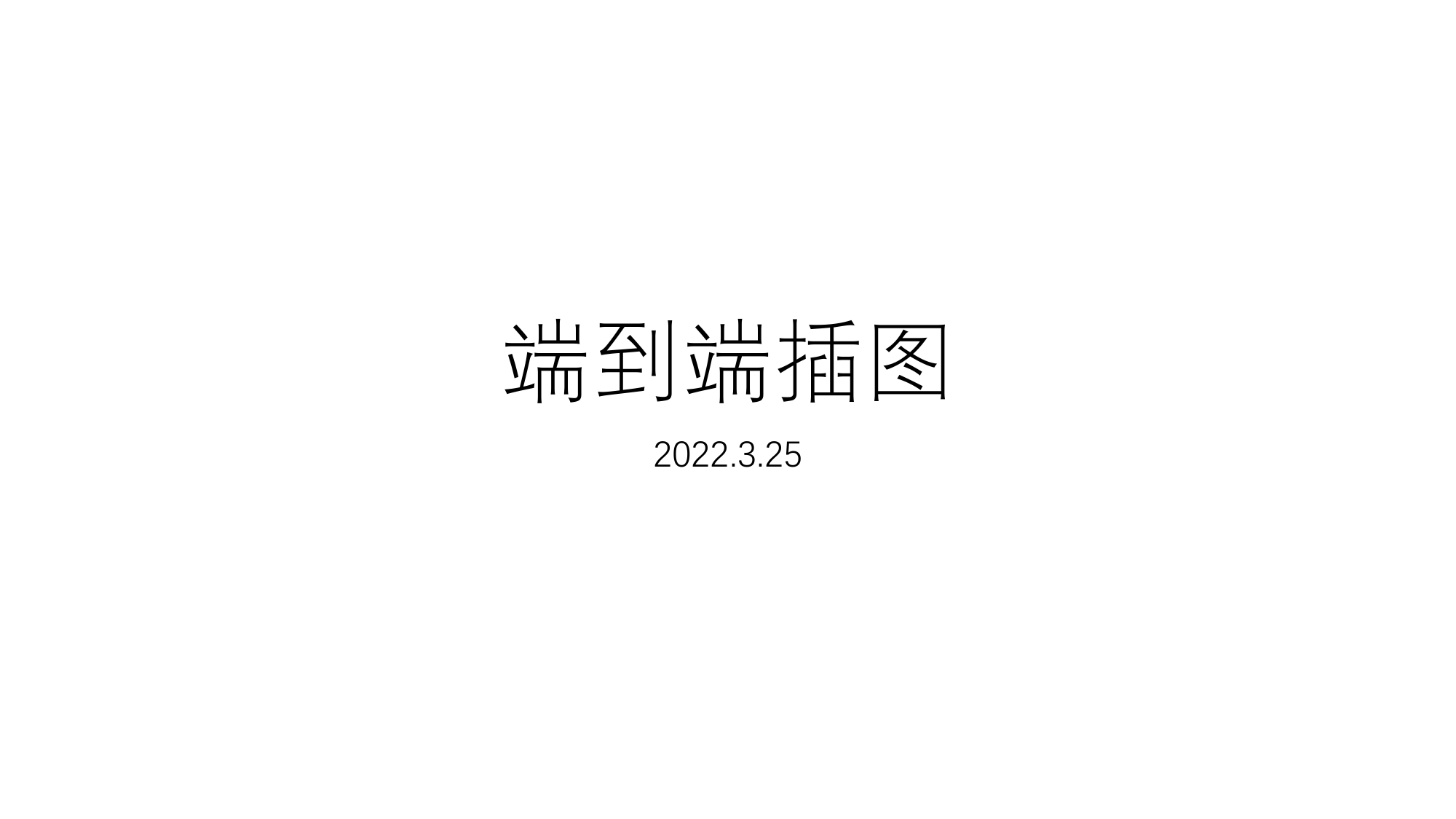}
        \caption{Separate transform and entropy estimation module in \cite{nakanishi2018neural}.}
        \label{fig:arch3}
    \end{subfigure}
    \hfill
    \begin{subfigure}[b]{1\columnwidth}
        \centering
        \includegraphics[width=0.5\textwidth]{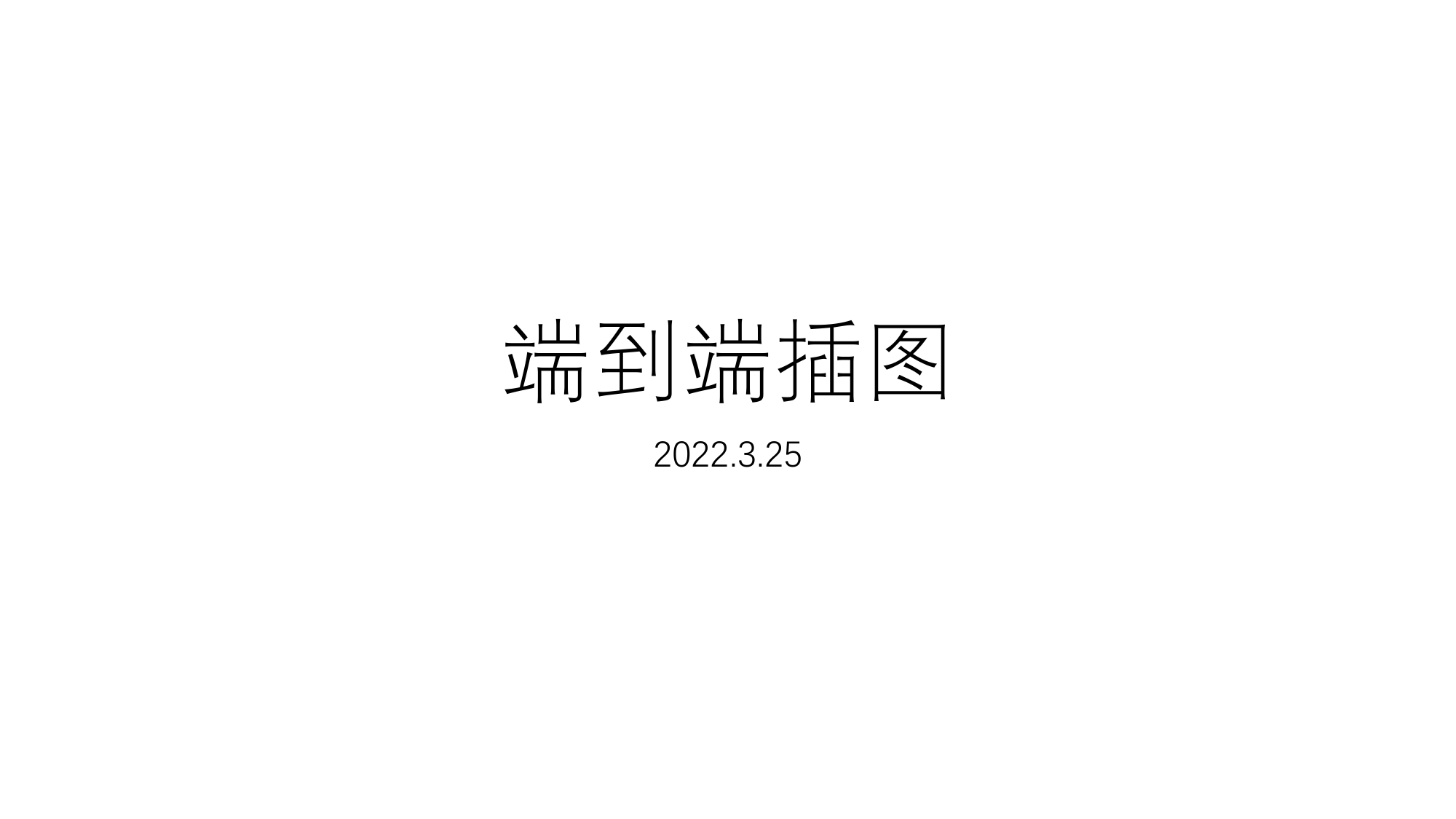}
        \caption{Proposed method with the frequency oriented transform and separate entropy estimation modules.}
        \label{fig:arch4}
    \end{subfigure}
    \caption{Comparison of the transform module used in end-to-end optimized image compression models. The unfilled rectangle represents an intermediate feature and the slashed rectangle means the feature used for entropy coding. The length of the rectangle represents its relative spatial size.}
\label{fig:archs}
\end{figure*}



In this paper, we propose an interpretable transform originated from the aspect of frequency decomposition, as illustrated in Fig.~\ref{fig:frequency_display}(a). 
Better transform results in promising spatial redundancy elimination \cite{rippel2017real,nakanishi2018neural}. Appropriate transform, numerical quantization, and data probability distribution estimation \cite{balle2016end,minnen2018joint,cheng2020image} all together result in a final compact bit-stream produced by entropy coding.
Moreover, frequency-oriented transform offers the interpretability.

First, we review the past works on the designs of the transform part in representative learning-based compression models in Fig.~\ref{fig:archs}. Fig.~\ref{fig:archs}(a) shows the baseline transform network architecture that the model only considers spatial correlation and spatial redundancy. Fig.~\ref{fig:archs}(b) takes multiple spatial scales into consideration but only encodes features in main stream, neglecting different data distribution in different scales. Fig.~\ref{fig:archs}(c) takes a step further by applying separate entropy estimation for different scales. However, the design in Fig.~\ref{fig:archs}(c) transforms one feature stream into multiple ones while neglects the correlation among different spatial scales and thus contain spatial-level data redundancy.

Based on these observations, we propose an interpretable convolution neural network (CNN)-based transform, namely frequency-oriented transform as illustrated in Fig.~\ref{fig:archs}(d) which decomposes image signal into learned frequency domains. We use separate symbol probability estimation modules to estimate the probability distribution in each frequency domain, considering their independent distribution property. The frequency-aware fusion module is proposed on the decoder side to support scalable coding and the attention module is adopted to capture the non-local spatial correlation.

Experimental results are analyzed to prove how the proposed frequency-oriented transform can efficiently reduce redundancies in both spatial and frequency domains.
Also, the scalable coding mechanism is supported by transmitting selected frequency components with our proposed model, considering strict bandwidth restrictions. In addition, to explore frequency-oriented compressed features' effectiveness in cognitive tasks, we further explore how arbitrary frequency components may impact downstream visual analysis tasks' results.

The contributions of this paper are summarized as follows:

\begin{itemize}

\item By analyzing compression degradation from the frequency aspect and reviewing past transforms in learning-based compression models, we propose an end-to-end image compression model with the frequency-oriented transform to disentangle the original image signal into the frequency domain and further eliminate signal redundancy in the target domain.

\item The frequency-oriented transform is proposed which is friendly for human interpretability. Both quantitative and qualitative analysis are conducted to verify the proposed method satisfies the human vision system from the aspect of frequency awareness. 

\item Extensive experiments conducted on different datasets (i.e., Kodak \cite{kodak} and CLIC2020 \cite{clic2020} professional test datasets) show that our model performance surpasses all traditional codecs (e.g., JPEG, H.265/HEVC, and H.266/VVC) on both datasets on MS-SSIM metric. Visual quality analysis is conducted to prove the proposed model effectively preserves semantic-related information.


\item Visual analysis performance on reconstructed images is examined with object detection and segmentation tasks. Experiments show that the analysis performance results better on the reconstructed images from the proposed compression model than from the codec H.266/VVC.

\end{itemize}

A preliminary version of this manuscript has been published in DCC 2022 \cite{Yuefeng2022Interpretable} and this manuscript provides the more details of the model and extensive visual analysis on the reconstructed images. The remaining of this paper is organized as follows. Section~\ref{sec:related} analyses the related work. In Section~\ref{sec:proposed}, we propose the end-to-end optimized image compression model with the spatial sampling, frequency-oriented transform, quantization and entropy estimation, and frequency-aware fusion. In Section~\ref{sec:implementation_details}, we display implementation details of the proposed model. Section~\ref{sec:experiments} shows experimental results and and gives frequency analysis. The visual analysis on multiple cognitive tasks is displayed in Section~\ref{sec:visual} and concluding remarks are given in Section~\ref{sec:conclusion}.

\section{Related Work}
\label{sec:related}

The proposed end-to-end optimized image compression model focuses on the frequency-oriented transform. We now discuss previous works related to multi-scale representation learning, frequency decomposition, interpretable machine learning, and advances in scalable coding.

\subsection{Multi-Scale Representation Learning}
In the deep learning era, multi-scale representation learning is widely applied because of its robustness and generalization. Inception series \cite{szegedy2017inception} attempt to utilize multiple branches with different spatial resolutions.
Some works explore multi-scale structures to further eliminate spatial redundancy existing in the feature domain \cite{rippel2017real,nakanishi2018neural}. Image Laplace pyramid is introduced in learning-based compression \cite{rippel2017real} but they do not consider the distribution difference of different resolution features as illustrated in Fig.~\ref{fig:archs}(b), and in Fig.~\ref{fig:archs}(c) \cite{nakanishi2018neural} considers this but leaves spatial redundancy.

Pyramid methods \cite{Burt1983TheLP,Adelson1984PYRAMIDMI} are used in various image processing tasks including image compression for a long history. Image can be decomposed into a set of spatial frequency bandpass component images, retaining both spatial localization as well as localization in the spatial-frequency domain. Within machine learning, variations of computer vision areas have introduced the idea of the spatial pyramid \cite{Ranjan2017OpticalFE,Li2021MultiScaleSI}.

\subsection{Frequency Decomposition}

Images can be transformed into a series of low-to-high frequency components. This idea is adopted in traditional image compression methods (e.g., wavelet \cite{antonini1992image} and DCT \cite{watson1994image} transform) and finally integrated into codecs (e.g, JPEG2000). 

With the rapidly growing usage of CNN in image processing and analysis, octave convolution \cite{chen2019drop} is proposed to substitute vanilla CNNs. However, octave convolution does not explore the relationship between each frequency component and mainly focuses on reducing channel-wise redundancy. In the super-resolution area, Omni-frequency \cite{li2020learning} is proposed to use CNNs with large strides to obtain low-frequency components and then derive higher frequency components from them. Recent research further employs the frequency-splitting idea of octave convolution and adopts it in the end-to-end compression model by splitting the original image into low-frequency signal and high-frequency signal \cite{akbari2020generalized}. While those methods lack scalability, our proposed method supports arbitrary split granularity with proposed transform and fusion modules.

\subsection{Interpretable Machine Learning}

Inscrutable black-box models suffer the problem of troubleshooting and are hard for humans to understand. Interpretability is crucial to trust for AI models \cite{Markus2021TheRO}. Supervised disentanglement of neural networks attempts to disentangle neurons with prior knowledge\cite{Rudin2021InterpretableML}. Suppose neuron $n$ in network layer $l$ is assigned concept $c$, then disentanglement constraint is that all signals about a specific concept $c$ in layer $l$ will only pass through assigned neuron $n$. In our case, as shown in Fig.~\ref{fig:frequency_display}, the original image signal will be disentangled into three frequency-oriented latents (visualized by their corresponding reconstructed images). To be noted, our human-interpretable concept used here is frequency decomposition which is implemented in an unsupervised way; details will be illustrated in Chapter \ref{sec:proposed}.

\subsection{Scalable Coding}

Most of today's learned end-to-end coding models need to be retrained for various bit-rate (i.e., each specific bit rate needs a corresponding model). Coding models which support scalable coding are proposed to solve this problem. Neural network structures are studied how to compress the image into residual information \cite{44844,jia2019layered}. The others explore extending the traditional Rate-Distortion (R-D) optimization strategy with the auxiliary gradient from training to dynamically allocate bit rates \cite{choi2019variable}.
Our approach follows the scalable encoding idea of decomposing input images from a frequency-oriented decomposition approach.

\section{Proposed Method}
\label{sec:proposed}

\begin{figure*}[t]
\begin{center}
   \includegraphics[width=0.95\linewidth]{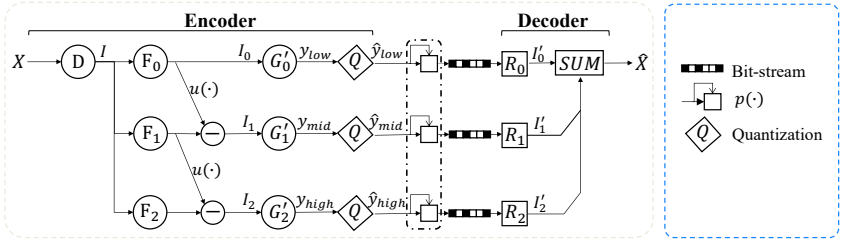}
\end{center}
\caption{Overview architecture of our proposed compression model. \textit{Q} is the quantization and \textit{SUM} represents the pixel-wise sum. The frequency-oriented transform decomposes the input image into frequency non-overlapping components, i.e., $y_{low}, y_{mid}, y_{high}$. The frequency-aware fusion module is designed for frequency selection.}

\label{fig:overview-architecture}
\end{figure*}

The proposed model consisting of spatial sampling (Section \ref{sec:Spatial Sampling}), frequency-oriented transform (Section \ref{sec:Frequency-Oriented Transform}), quantization and entropy estimation (Section \ref{sec:Quantization and Entropy Estimation}), and frequency-aware fusion (Section \ref{sec:Frequency-Aware Fusion}). In Section \ref{sec:Optimization Target}, we describe the optimization target for the proposed model, and Fig.~\ref{fig:overview-architecture} provides an overview of the proposed model.

\subsection{Spatial Sampling}
\label{sec:Spatial Sampling}

Suppose that $X$ is the original image with dimensions $m \times n$ that are powers of 2 and $\hat{X}$ is the corresponding reconstructed image after decoding. Let $\mathcal{C}$ represents convolution operations. $\mathcal{C}_{\downarrow2}(\cdot)$ refers to $3 \times 3$ convolution with stride=2 and $\mathcal{C}_{\downarrow2}(X)$ has size $m/2 \times n/2$. $u(\cdot)$ is the bilinear upsampling operation with a factor of 2.

To transform the original image $X$ into the compact feature space aiming to remove pixel-to-pixel correlations, we first downsample the input image from the spatial view by two consecutive convolution layers as:
\begin{equation}
I = \mathcal{C}_{\downarrow2} (\mathcal{C}_{\downarrow2}(X)),
\label{eqa:downsample}
\end{equation}
where $I$ is the spatial-downsampled feature with size of $m/4 \times n/4$. We denote the size of $I$ as $m' \times n'$ without the loss of generality.

\subsection{Frequency-Oriented Transform}
\label{sec:Frequency-Oriented Transform}

Inspired by the Laplace pyramid transform \cite{Burt1983TheLP}, we propose the transform model as Fig.~\ref{fig:arch4} to transform the images into a frequency domain with the help of scale variations. Meanwhile, we utilize separate entropy estimation models for each frequency split, based on the hypothesis that their distributions are independent.

Let $\{F_0, ..., F_K\}$ denote a set of convolutional neural networks composed of $\mathcal{C}$ operation and $I_0$ represents for the first layer: $I_0 = F_0(I)$. Then the immediate result is subtracted recursively layer by layer and the target feature of the $k$-th layer is as:
\begin{equation}
\begin{aligned}
    I_k = F_k(I) - u(F_{k-1}(I))
\end{aligned}
\end{equation}

Each level in the pyramid represents a frequency bandpass from low to high frequency. To eliminate spatial information dependency in each layer, a set of networks $\{F_0', ..., F_K'\}$ is applied with the spatial change as $\{F_0, ..., F_K\}$, to unify the feature size of each frequency split. Then intermediate features can be calculated as:
\begin{equation}
\begin{aligned}
    y_k = F_k'(I_k)
\end{aligned}
\end{equation}
where $y_k$ should have the same spatial resolution as $m'/2^{K-1} \times n'/2^{K-1}$.

In practice, we use a 3-level pyramid ($K = 2$) and $y_0, y_1, y_2$ can be written as $y_{high}, y_{mid}, y_{low}$ representing high-, middle- and low-frequency features, respectively.
The frequency-oriented transform $\{F_0, ..., F_K\}$ and $\{F_0', ..., F_K'\}$ can be detailedly formulated as:
\begin{equation}
\begin{aligned}
y_{high} &= \mathcal{C}_{\downarrow2}(\mathcal{C}_{\downarrow2}(I)), \\
y_{mid} &= \mathcal{C}_{\downarrow2}(I) - u(y_{high}), \\
y_{low} &= \mathcal{C}_(I) - u(y_{mid}).
\end{aligned}
\label{eqa:frequency-pyramid}
\end{equation}



\subsection{Quantization and Entropy Estimation}
\label{sec:Quantization and Entropy Estimation}

Signal entropy can be reduced by quantizing intermediate feature values which is the key step in lossy compression compared with lossless one. After decomposing the input image into frequency-oriented features, we conduct quantization and entropy estimation on these features.

\subsubsection{Quantization}
To promise the calculation of gradients in the backward broadcasting, the quantization process is replaced by adding a uniform noise $\mathcal{U}\left(-\frac{1}{2}, \frac{1}{2}\right)$ during the training phase. And round-based quantization is adopted during the inference step. Here we set $Q(\cdot)$ as the quantization operation which output is noted as $\hat{y} = Q(y)$. In Fig.~\ref{fig:overview-architecture}, $\hat{y}_{high}, \hat{y}_{mid}, \hat{y}_{low}$ are quantized features of high-, middle- and low-frequency, respectively.

\subsubsection{Entropy Estimation}

In the entropy estimation process, we use a hyperprior \cite{balle2018variational} to further eliminate spatial dependencies among latent features. The probability estimation is as:

\begin{equation}
p_{\hat{y}_{i} \mid \hat{z}_{i}}\left(\hat{y}_{i} \mid \hat{z}_{i}\right), i \in\{high,  mid, l o w\},
\end{equation}
where $\hat{z}_i$ is the  hyperprior and $p_{\hat{y}_i \mid \hat{z}_i}(\hat{y}_i \mid \hat{z}_i)$ is the estimated distribution of each frequency split conditioned on corresponding $\hat{z}_i$. We adopt range Asymmetric Numeral Systems (ANS)\cite{Duda2013AsymmetricNS} as the entropy encoder to code the frequency features into the actual bit-streams.

\subsection{Frequency-Aware Fusion}
\label{sec:Frequency-Aware Fusion}

Frequency-aware fusion targets to combine latent features of each frequency split together. We first unify channel dimension number by operations $\{R_0, ..., R_K\}$, as:

\begin{equation}
\begin{aligned}
I_{0}' &= R_0(\hat{y}_{low}), \\
I_{1}' &= R_1(\hat{y}_{mid}), \\
I_{2}' &= R_2(\hat{y}_{high}),
\end{aligned}
\label{eqa:reconstructed-pyramid}
\end{equation}
where $I_0', I_1', I_2'$ are the reconstructed results for each frequency split.

We take a point-wise sum up (denoted as \textit{SUM} in Fig.~\ref{fig:overview-architecture}) for frequency feature fusion at the decoder side. The reconstructed image $\hat{X}$ can be represented as:
\begin{equation}
\hat{X} = \textit{SUM}(I_0', I_1', I_2').
\end{equation}
With the \textit{SUM} operation, we could manually transmit the features with arbitrary combinations, i.e., transmitting low-frequency band combinations under limited bandwidth and high-frequency ones under high bit-rate coding scenarios. 

\begin{figure}[t]
\begin{center}
  \includegraphics[width=0.6\linewidth]{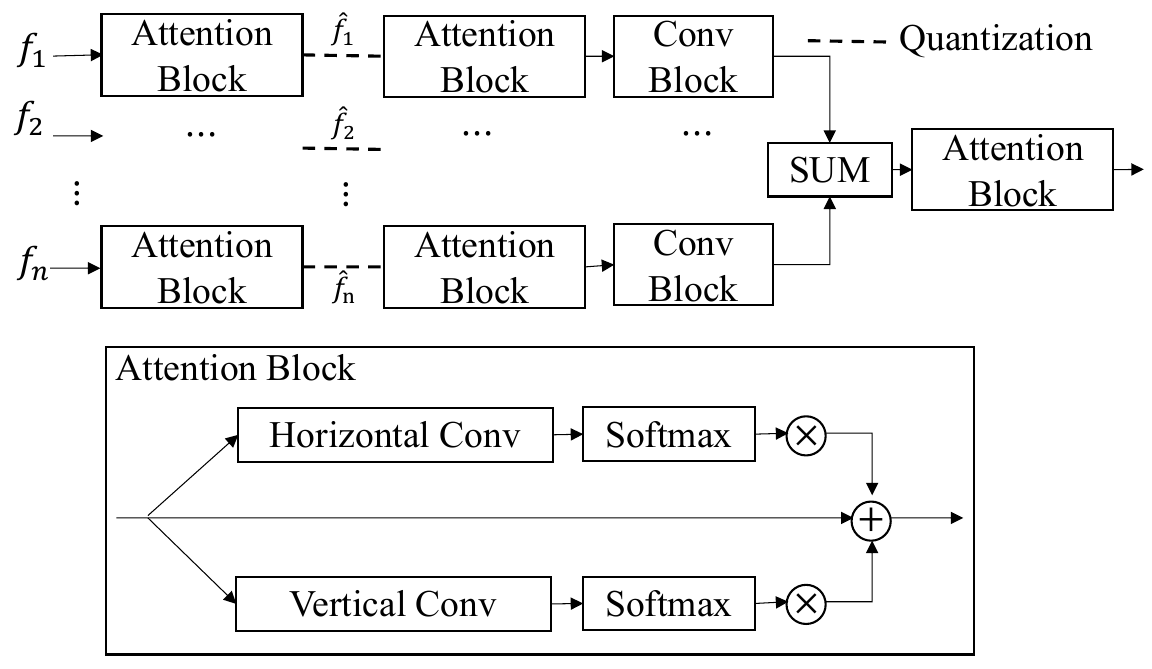}
\end{center}
\caption{Illustration of frequency-aware fusion. Selective frequency components are aggregated together at the decoder side by \textit{SUM} operation. $f$ represents the  feature matrix.}
\label{fig:fusion-module}
\end{figure}

Details of the frequency-aware fusion is shown in Fig.~\ref{fig:fusion-module} that attention blocks are added after convolution operations. Our attention module references the design of a criss-cross attention block \cite{huang2019ccnet}. Long-distance reliances from both horizontal and vertical orientations are taken into consideration, thus in the proposed attention module horizon-aware attention and vertical-aware attention are combined as shown at the bottom of Fig.~\ref{fig:fusion-module}.

\begin{figure*}[t]
  \centering
  \includegraphics[width=0.49\linewidth]{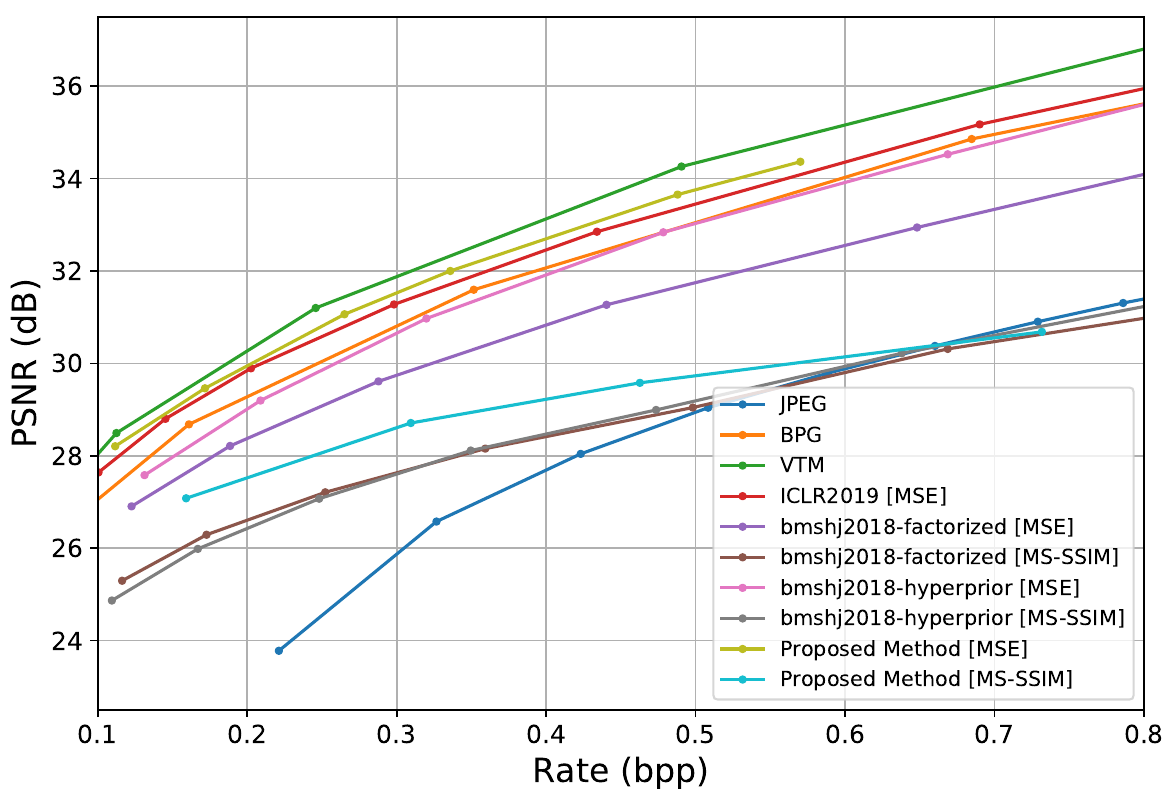}
  \includegraphics[width=0.49\linewidth]{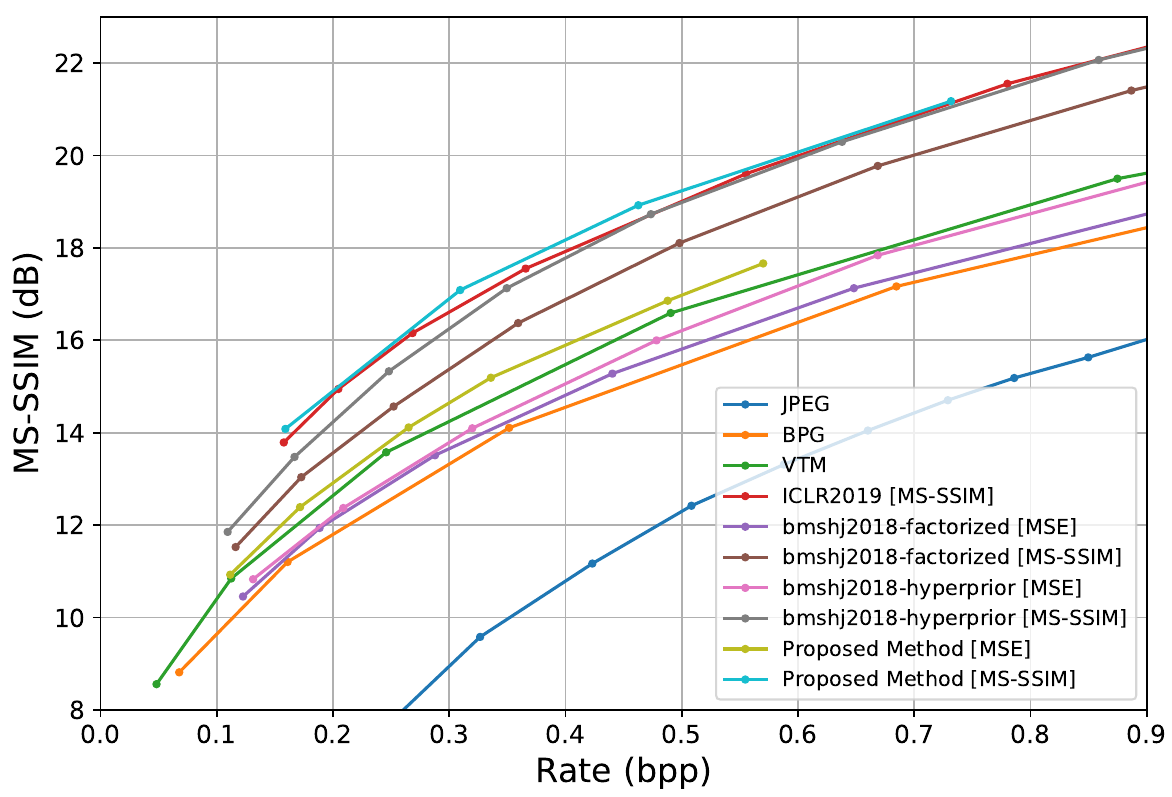}
  \caption{Compression performance evaluation on Kodak dataset.}
  \label{fig:RD-curve-kodak}
  
  \includegraphics[width=0.49\linewidth]{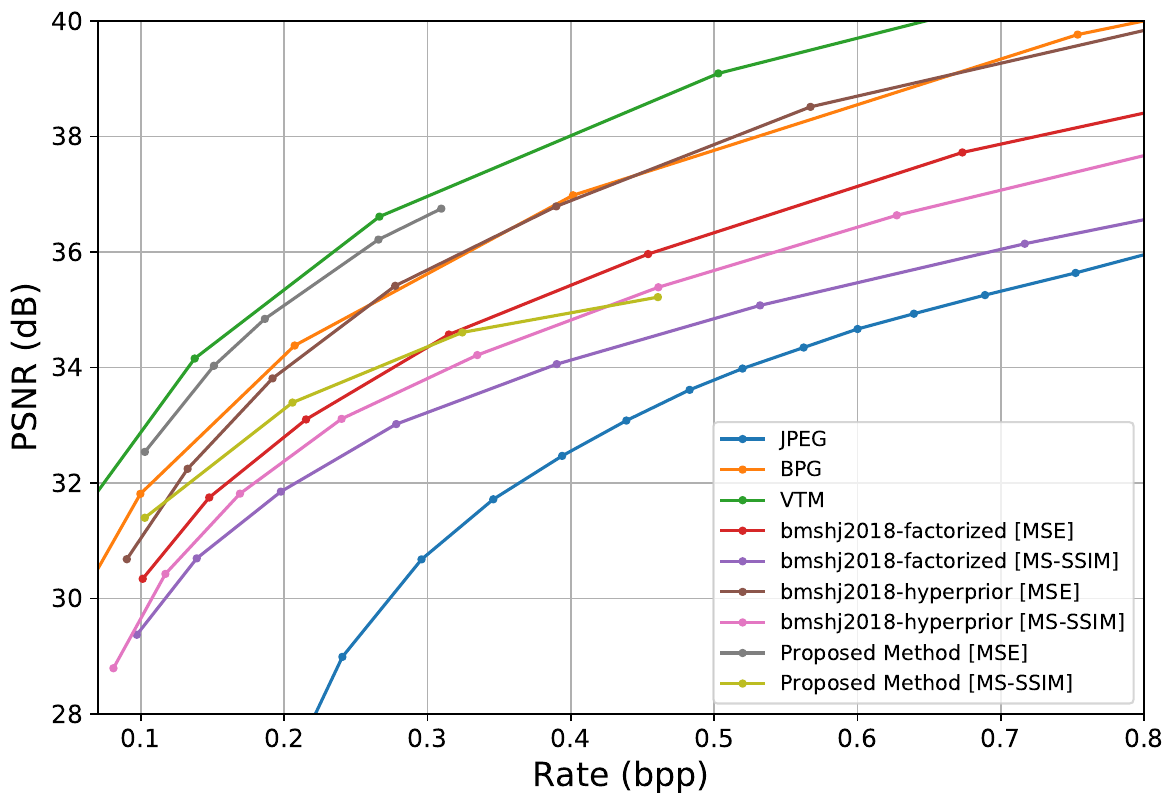}
  \includegraphics[width=0.49\linewidth]{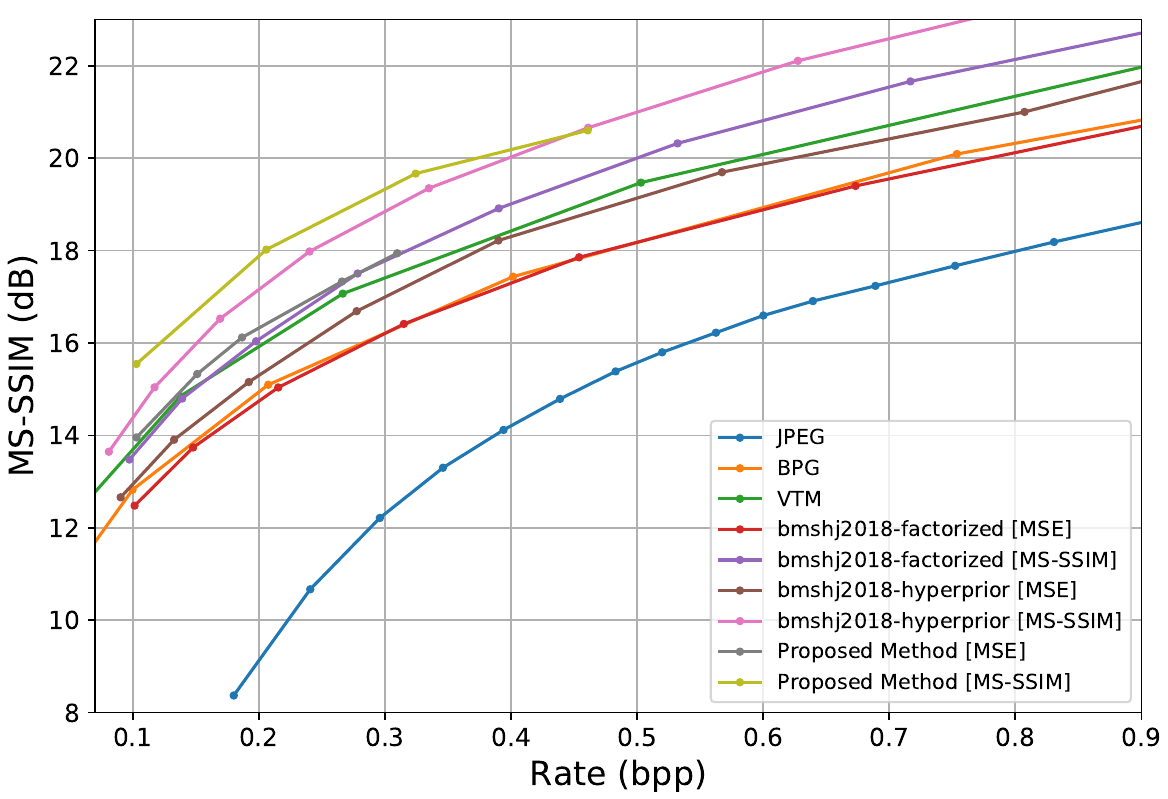}
  \caption{Compression performance evaluation on CLIC2020 professional test dataset.}
  
  \label{fig:RD-curve-clic}
\end{figure*}


\subsection{Optimization Target}
\label{sec:Optimization Target}

Our proposed compression model is trained in an end-to-end way by optimizing a Lagrangian multiplier-based rate-distortion optimization as following:
\begin{equation}
\begin{aligned}
\mathcal{L} &=\sum_{k}\left[\mathcal{R}\left(\hat{y}_{k}\right)+\mathcal{R}\left(\hat{z}_{k}\right)\right]+\lambda \cdot \mathcal{D}(X, \hat{X}) \\
&=\sum_{k}\{\mathbb{E}\left[-\log \left(p_{\hat{y}_{k} \mid \hat{z}_{k}}\left(\hat{y}_{k} \mid \hat{z}_{k}\right)\right)\right] \\ &+\mathbb{E}\left[-\log \left(p_{\hat{z}_{k} \mid \psi_{k}}\left(\hat{z}_{k} \mid \psi_{k}\right)\right)\right]\} +\lambda \cdot \mathcal{D}(X, \hat{X}),  k \in K,
\end{aligned}
\label{eqa:loss-function}
\end{equation}

\noindent where $\psi_k$ is the factorized density model following \cite{balle2018variational} and hyper-parameter $\lambda$ balances the trade-off between rate and distortion. $K$ are positive integers. $\mathcal{D}(X, \hat{X})$ represents the distortion metric between the original images and the reconstructed ones and we discuss it detailedly in Section \ref{sec:implementation_details}.

\section{Implementation Details}
\label{sec:implementation_details}

We first display training details of the proposed method (Section \ref{sec:Training Details}). Next, we display different datasets and metrics (Section \ref{sec:Evaluation}) and both traditional and learning-based codecs (Section \ref{sec:Compared Methods}) for evaluating the effect of the proposed method.

\subsection{Training Details}
\label{sec:Training Details}
We use the official training split from the Vimeo-90k triplet dataset \cite{xue2019video} for training and randomly crop them  with the size of $256 \times 256$ pixels. We use Adam optimizer \cite{kingma2014adam} with a mini-batch size of 32 and the initial learning rate is 1e-4 which is divided by 2 when the evaluation loss arrives at a plateau. We train each model for a total of 450k iterations.

Two distortion metrics are used: mean square error (MSE) and multiscale structural similarity (MS-SSIM). When using MSE, $\lambda$ is set as \{0.0035, 0.0067, 0.01, 0.025\} and $\mathcal{D}(x, \hat{x}) = MSE(x, \hat{x}) $. For MS-SSIM, $\lambda$ belongs to \{4, 16, 40, 120\} and we have $\mathcal{D}(x, \hat{x}) = 1 - MS\_SSIM(x, \hat{x})$. We present the RD curves to demonstrate the coding efficiency as shown in Fig.~\ref{fig:RD-curve-kodak} and Fig.~\ref{fig:RD-curve-clic} which are analyzed in Section \ref{sec:experiments}.

\subsection{Evaluation}
\label{sec:Evaluation}

We first present the datasets (Section \ref{sec:Evaluation Datasets}) and then metrics (Section \ref{sec:Metrics}) for the evaluation.

\subsubsection{Evaluation Datasets}
\label{sec:Evaluation Datasets}

\textbf{Kodak:}
Kodak \cite{kodak} is a widely used testing dataset for evaluating image compression performance. The Kodak dataset consists of 24 lossless images with the resolution $512 \times 768$ pixels. Kodak dataset has various contents and textures, which are commonly used in evaluating image compression methods.

\noindent \textbf{CLIC2020 Professional Dataset:}
Recently, the Challenge on Learned Image Compression (CLIC)  \cite{clic2020} has caught much attention in the area of learning-based image compression. CLIC2020 professional test dataset provides high-quality images with an average resolution of $1803 \times 1175$ pixels, which contains 250 images for test split.

\subsubsection{Metrics}
\label{sec:Metrics}
To evaluate the rate-distortion (RD) performance, we compare the methods using both PSNR and MS-SSIM metrics. MS-SSIM value is described in decibels as -10 log10 (1 - MS-SSIM). The comparison results are thoroughly discussed in Section \ref{sec:Objective Performance}.

\subsection{Compared Methods}
\label{sec:Compared Methods}

Both traditional codecs (Section \ref{sec:Traditional Codecs}) and learning-based codecs (Section \ref{sec:Learning-based Codecs}) are chosen as the compared methods to evaluate the effect of the proposed method.

\subsubsection{Traditional Codecs}
\label{sec:Traditional Codecs}
We compare the test results with traditional codecs such as JPEG, H.265/HEVC, and next generation H.266/VVC standard. We use the PIL library \cite{PIL} for JPEG. We use the BPG software \cite{bellard2014bpg} for H.265/HEVC and test software VTM version 11.0 \cite{VTM} for H.266/VVC which has all-intra mode with 8-bit YCbCr 4:4:4 as the configuration.

\subsubsection{Learning-based Codecs}
\label{sec:Learning-based Codecs}

We compare our method with state-of-the-art learning-based codecs, including bmshj2018-factorized \cite{balle2018variational}, bmshj2018-hyperprior \cite{balle2018variational}, mbt2018-mean \cite{minnen2018joint}, and ICLR2019 \cite{Lee2019Context}. Each of these methods employs two models that optimize for both PSNR and MS-SSIM metrics during the training process, ensuring a comprehensive evaluation of performance. The implementation of the compared methods is credited to the CompressAI library \cite{Begaint2020CompressAIAP}.

\begin{figure}[t]%
\centering

\includegraphics[width=1\linewidth]{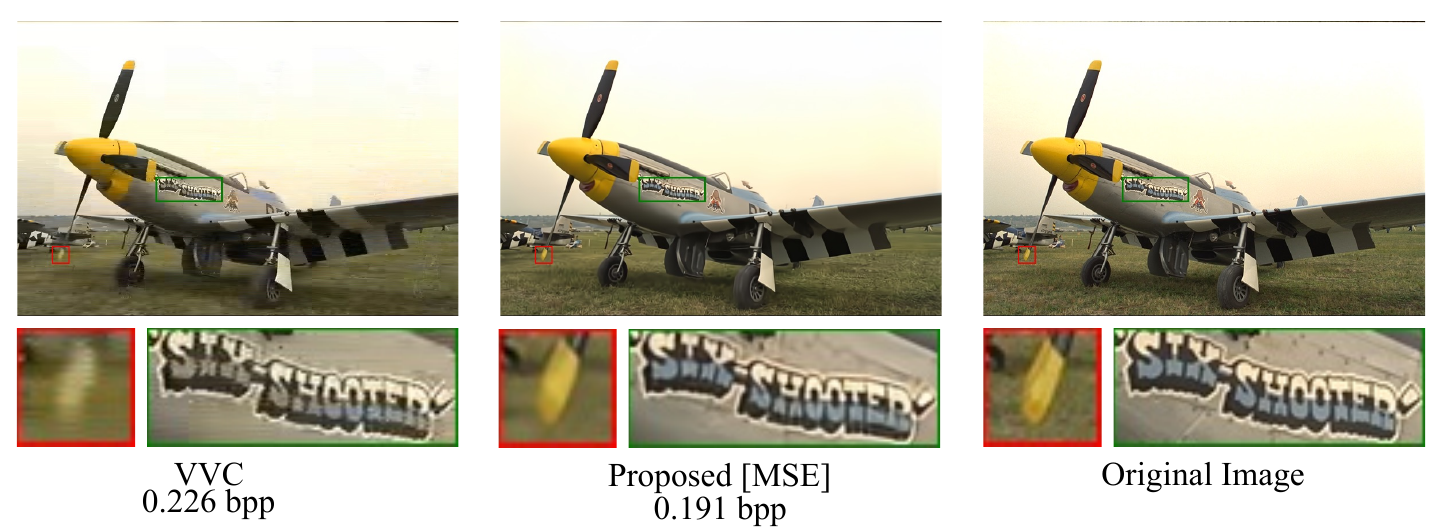}

\caption{Visualization of reconstructed images \textit{kodim20} from Kodak dataset. Our proposed model is optimized by MSE loss and $\lambda$ is set as $0.01$.}

\label{fig:visual_kodim20}
\end{figure}

\begin{figure}[t]%
\centering

\includegraphics[width=1\linewidth]{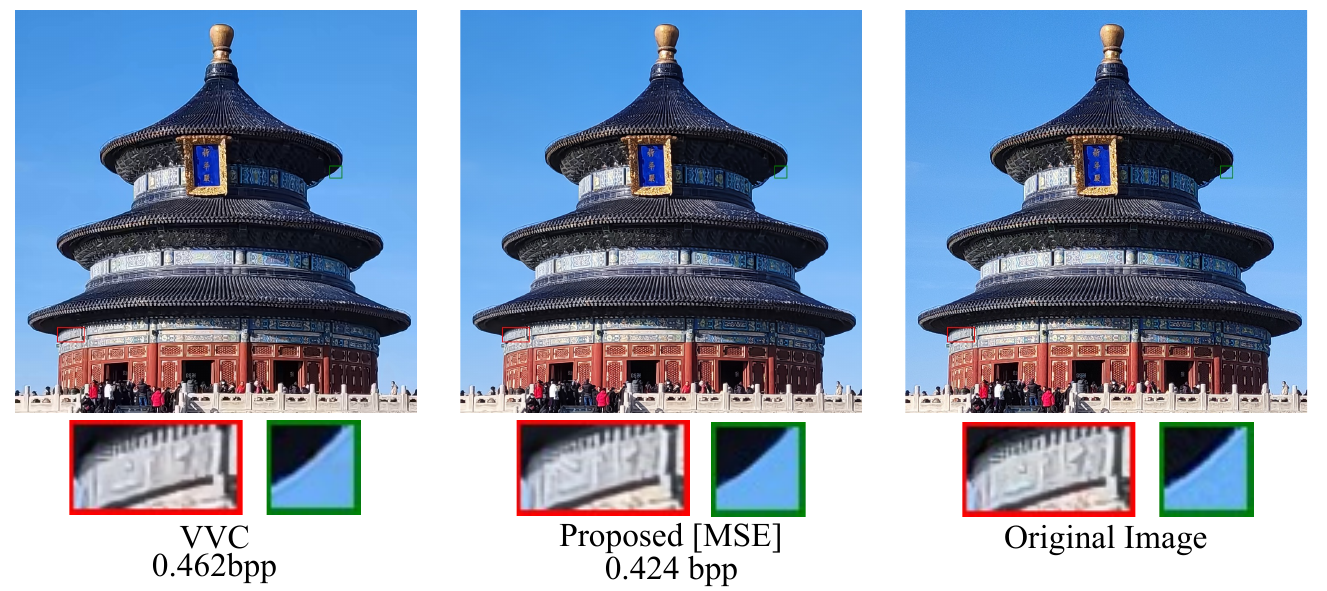}

\caption{Visualization of reconstructed images \textit{Qiniandian} captured by the camera. Our proposed model is optimized by MSE loss and $\lambda$ is set as $0.01$.}

\label{fig:visual_qiniandian}
\end{figure}

\begin{table}[t]
\centering

\caption{BD-Rate evaluation on different methods with PSNR as the distortion metric. Codec BPG is set as the anchor method.}

\begin{tabular}{@{}lll@{}}
\toprule
Methods & Kodak & CLIC2020 \\ \midrule
ICLR2018\_Factorized & 30.59\% & 41.67\% \\
ICLR2018\_Hyperprior & 4.80\% & 1.53\% \\
ICLR2019 & -5.03\% & 38.89\% \\
Ours & -12.75\% & -17.12\% \\
VVC & -19.31\% & -24.50 \% \\
BPG & 0.00 & 0.00 \\
JPEG & 116.24\% & 140.04\% \\ \bottomrule
\end{tabular}

\label{table:BD-rate}
\end{table}

\begin{table}[t]
\centering

\caption{Evaluation of image objective qualities on different combinations of frequency splits. $\lambda$ for the compression model is set as $0.01$.}

\begin{tabular}{@{}llllll@{}}
\toprule
\multicolumn{3}{c}{Frequency} & \multicolumn{2}{c}{Metrics} & \multirow{2}{*}{\begin{tabular}[c]{@{}l@{}}Rate \\ (bpp)\end{tabular}} \\ \cmidrule(r){1-3} \cmidrule(r){4-5}
Low & Middle & High & PSNR (dB) & MS-SSIM &  \\ \midrule
$\checkmark$ & \xmark & \xmark & 18.19 & 0.881 & 0.172 \\
$\checkmark$ & $\checkmark$ & \xmark & 23.36 & 0.930 & 0.193 \\
$\checkmark$ & \xmark & $\checkmark$ & 22.03 & 0.931 & 0.315 \\
$\checkmark$ & $\checkmark$ & $\checkmark$ & 32.00 & 0.970 & 0.336 \\ \bottomrule
\end{tabular}

\label{table:frequency-split-metrics}
\end{table}

\section{Experiments}
\label{sec:experiments}

In this section, we display the experiments results of the proposed method of both objective performance (Section \ref{sec:Objective Performance}) and subjective performance (Section \ref{sec:Subjective Performance}). In Section \ref{sec:Ablation Study}, ablation study is conducted to explore the effect of the attention module and model computation complexity. In order to evaluate the frequency-oriented transform, we analysis the frequency splits in Section \ref{sec:Frequency Analysis}.

\subsection{Objective Performance}
\label{sec:Objective Performance}

Both the rate-distortion curve (Section \ref{sec:Rate-Distortion Curve}) and the BD-rate with PSNR (Section \ref{sec:BD-Rate with PSNR}) results are presented to illustrate the objective performance comparison between the compared codecs and the proposed method.

\subsubsection{Rate-Distortion Curve}
\label{sec:Rate-Distortion Curve}

Rate-distortion curves of the proposed method and competitive methods are in Fig.~\ref{fig:RD-curve-kodak} and Fig.~\ref{fig:RD-curve-clic}. The proposed method achieves significant gains over traditional codecs including H.266/VVC in terms of MS-SSIM quality metric. The proposed model shows comparable performance with H.266/VVC on the PSNR metric, which will be further discussed in the following section.

\subsubsection{BD-Rate with PSNR}
\label{sec:BD-Rate with PSNR}

We evaluate rate-distortion performance on the Kodak dataset and CLIC2020 professional test dataset. As illustrated in Fig.~\ref{fig:RD-curve-kodak} and Fig.~\ref{fig:RD-curve-clic}, our method shows comparable results on the PSNR metric with VVC, meanwhile outperforming other traditional codecs (i.e. JPEG, JPEG2000, HEVC). On the MS-SSIM metric, our proposed method optimized by MS-SSIM loss performs better than all traditional codecs including VVC on both datasets.

To give a better illustration of the difference between our method and the others, we evaluate the 
Bjøntegaard Delta (BD)-rate with PSNR against anchor BPG on both datasets as shown in Table~\ref{table:BD-rate}. 
We adopt the bit-rate range as [0.4, 1.15] for the Kodak dataset and [0.3, 0.9] for the CLIC dataset following the setting in \cite{hu2021learning}. BD-PSNR gain of the proposed model is $12.75\%$ and $20.81\%$ on Kodak and CLIC datasets, respectively.

\subsection{Subjective Performance}
\label{sec:Subjective Performance}

The visual quality comparisons are provided in Fig.~\ref{fig:visual_kodim20} and Fig~\ref{fig:visual_qiniandian}, where images are compressed by the anchor (VTM 11.0) and the proposed method at a comparable bit rate. As shown in Fig.~\ref{fig:visual_kodim20}, the proposed method achieves better visual quality for the font shape and the object edges. It can be found that the anchor method suffers problems of blurriness (propeller in the red box) and color distortion (text in the green box) in the areas around the object edges , which can be eliminated by the proposed method. In the red box of Fig.~\ref{fig:visual_qiniandian}, it is observed that pattern details are lost in the decoded image by the anchor, whereas those details are reserved in the decoded image by the proposed method. Moreover, the original image in the green box of Fig.~\ref{fig:visual_qiniandian} captured by a digital camera has inherent ringing artifacts but the proposed method reduces those artifacts and provides a human-friendly reconstructed image. Those observations prove that the proposed image compression method can acquire human-friendly image details and improve the visual quality subjectively.

\subsection{Ablation Study}
\label{sec:Ablation Study}

Our ablation study includes - (i) how attention module effects (Section \ref{sec:Attention Module});(ii) discussing the computation complexity of the proposed method (Section \ref{sec:Computation Complexity}).

\subsubsection{Attention Module}
\label{sec:Attention Module}
We illustrate the effectiveness of the frequency-aware fusion module. Specifically, the proposed attention block in the frequency-aware fusion module is compared with the non-local attention block from \cite{cheng2020image} in Table \ref{table:attention_module}. Under the restriction that both comparison methods are trained following the same setting, the frequency-aware compression model with our proposed attention module reduces more than $50\%$ entropy loss, which means that the model effectively converges assisted by more accurate probability estimation.

Specifically, the proposed attention module shows a great advantage in module computation complexity and module parameter number compared with other attention modules used in the end-to-end image compression model, as displayed in Table \ref{table:attention_module}. In Table \ref{table:attention_module}, entropy loss denotes the mean average error between the estimated latents' distribution and the real data distribution. Rate loss is calculated as $\sum_i\left[\mathcal{R}\left(\hat{y}_{i}\right)+\mathcal{R}\left(\hat{z}_{i}\right)\right]/n_{pixels}, i \in \{low, mid, hi\}$, where $n_{pixels}$ is the image pixel number. The average loss is calculated as Eq.~(\ref{eqa:loss-function}).

\begin{table}[t]
\centering

\caption{Different attention modules' effect on model training loss. Rate, entropy and the average loss are compared. Here we display the model's results under $\lambda=0.01$.}
\label{table:attention_module}

\begin{tabular}{@{}lllll@{}}
\toprule
\multirow{2}{*}{\begin{tabular}[c]{@{}l@{}}Non-Local\\\end{tabular}}  & \multirow{2}{*}{\begin{tabular}[c]{@{}l@{}}Proposed\\\end{tabular}}  & \multicolumn{3}{c}{Loss} \\
 \cmidrule(r){3-5}
 & & Rate & Entropy  & Average \\ \midrule
\xmark & \xmark & 0.27 & 309.85 & 0.576 \\
$\checkmark$ & \xmark  & \textbf{0.26} & 434.65 & 0.577 \\
\xmark & $\checkmark$ & \textbf{0.26} & \textbf{213.22} &  \textbf{0.573} \\ \bottomrule
\end{tabular}

\end{table}

\begin{table}[t]
\centering
\caption{Attention module computation complexity and the number of model parameters when input vector size is [1, 64, 16, 16].}
\label{table:attention_param}
\begin{tabular}{@{}lll@{}}
\toprule
Attention Module & Macs & \#Params \\ \midrule
Proposed & \textbf{1.33M} & \textbf{5200} \\
Non-local \cite{Chen2021EndtoEndLI} & 4.26M & 16640 \\ 
AttentionBlock \cite{cheng2020image} & 21.71M & 84800 \\ \bottomrule
\end{tabular}

\end{table}

\begin{figure}
    \begin{subfigure}[b]{1\columnwidth}
        \centering
        \includegraphics[width=0.5\textwidth]{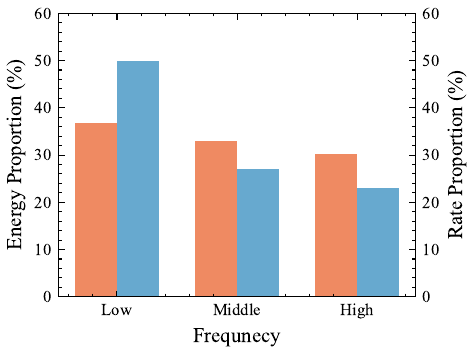}
        \caption{Original images' frequency proportion v.s frequency components' bit rate proportion of our proposed method.}
    \end{subfigure}

    \begin{subfigure}[b]{1\columnwidth}
        \centering
        \includegraphics[width=0.5\textwidth]{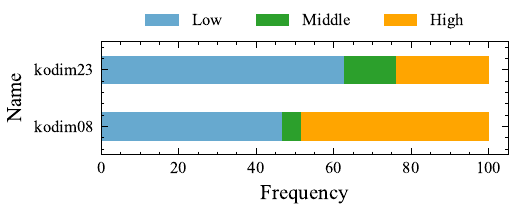}
        \caption{Two example images' bit rate proportion of each frequency component. The frequency band with Fourier transform is split equally in (a). }
    \end{subfigure}
    \caption{Illustration of the relationship between frequency energy distribution and bit-rates needed in data compression. $\lambda$ used in the proposed compression model is set as $0.0175$.}
\label{fig:frequency_analysis}
\end{figure}

\begin{figure*}[t]%
\centering
\subfloat[\centering Visualization of reconstructed results of \textit{kodim08} from Kodak dataset.]{
    {\includegraphics[width=1.0\linewidth]{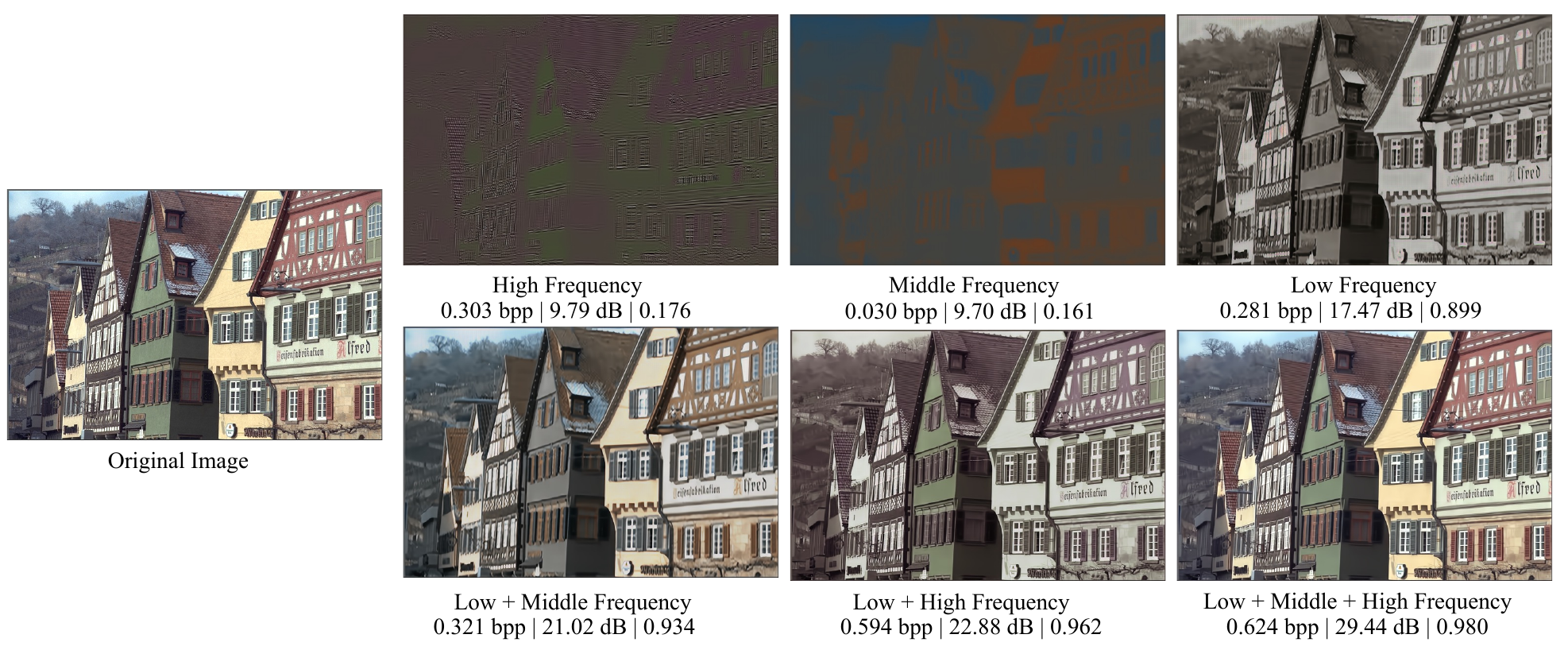} }
}

\subfloat[\centering Visualization of reconstructed results of \textit{kodim23} from Kodak dataset.]{
    {\includegraphics[width=1.0\linewidth]{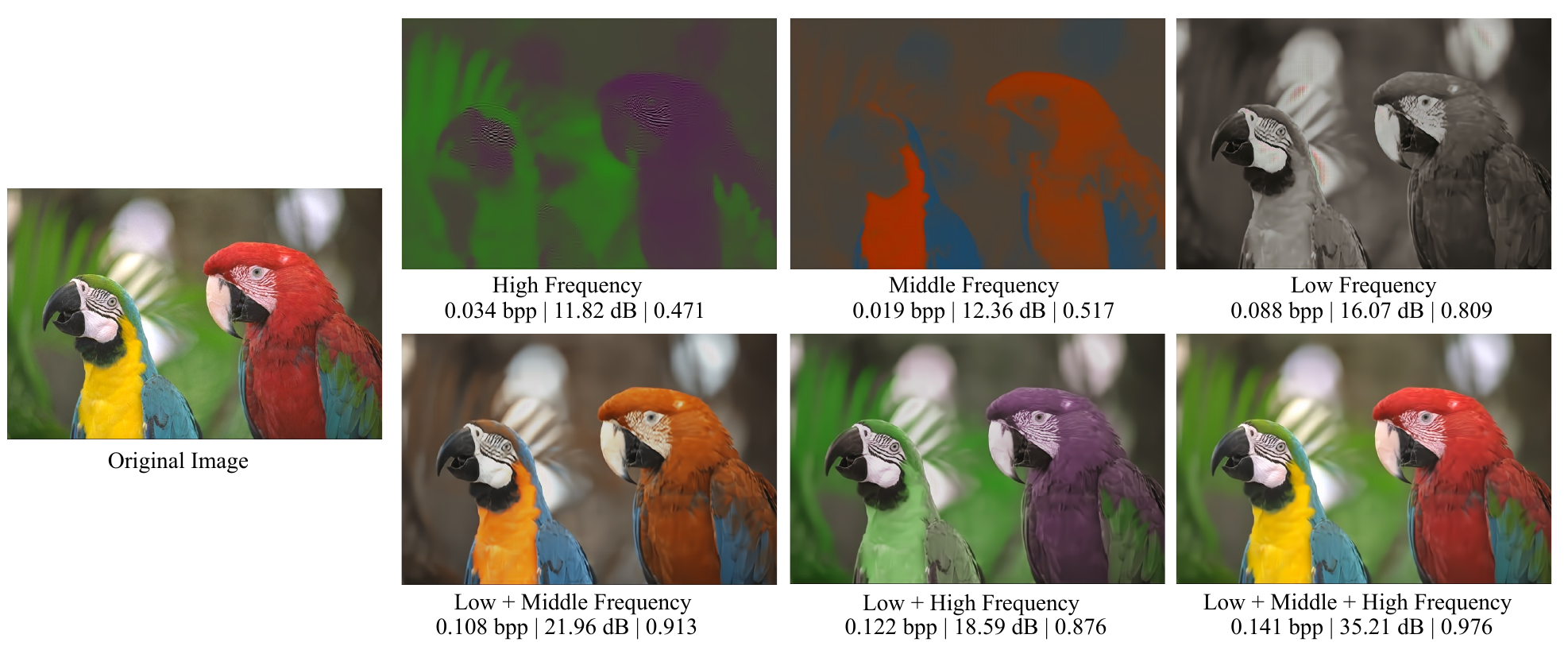} }
}
\caption{Visualization of reconstructed images and a few frequency-split combinations result from the Kodak dataset. Bit rate distribution of each frequency component is shown in Fig.~\ref{fig:frequency_analysis}(b). For each reconstructed image, we display its bit rate (i.e., in bit per pixel, bpp), PSNR (i.e., in dB), and MS-SSIM value under itself.}
\label{fig:08_23_display}
\end{figure*}

\begin{figure*}[ht]
    \begin{subfigure}[b]{1\columnwidth}
        \centering
        \includegraphics[width=0.6\textwidth]{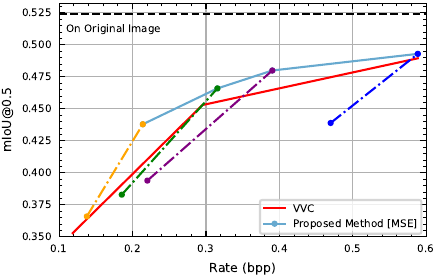}
        \caption{Detection task.}
    \end{subfigure}
    \hfill
    \begin{subfigure}[b]{1\columnwidth}
        \centering
        \includegraphics[width=0.6\textwidth]{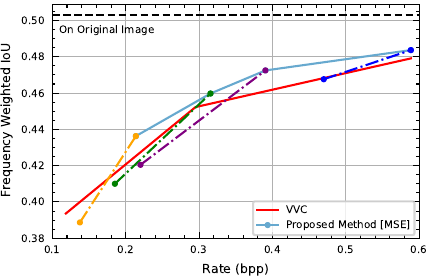}
        \caption{Segmentation task.}
    \end{subfigure}
    \caption{Comparison of different visual analysis tasks. Data points of the proposed model have different $\lambda$ settings that from left to right it is \{$0.0035, 0.0067, 0.01, 0.025$\}, sequentially. We display visual analysis results on different combinations of the frequency components, on the same dashed line, from low to high, inputs are in the set \{low+middle, low+middle+high\}.}
\label{fig:multiple-tasks}
\end{figure*}

\subsubsection{Computation Complexity}
\label{sec:Computation Complexity}

In Table \ref{table:attention_param}, we compare the parameter number of the proposed fusion module by replacing the attention module with Non-local \cite{Chen2021EndtoEndLI} and AttentionBlock \cite{cheng2020image} one. It shows that our proposed attention module saves $68.9 \%$ macs and $68.8\%$ \#params when compared with the Non-local \cite{Chen2021EndtoEndLI} module. The performance comparison is illustrated in Section \ref{sec:Attention Module}.

\subsection{Frequency Analysis}
\label{sec:Frequency Analysis}

We demonstrate the effect of each frequency split from both objective (Section \ref{sec:Objective Frequency Analysis}) and subjective aspects (Section \ref{sec:Subjective Frequency Analysis}).

\subsubsection{Objective Frequency Analysis}
\label{sec:Objective Frequency Analysis}

We illustrate each frequency component's effect on reconstruction images' objective quality on the Kodak dataset in Table \ref{table:frequency-split-metrics}. In each line, some of the frequency splits are masked to see their effect on the reconstructed image quality on both metrics. The results are consistent with the intuition that the more frequency splits are included, the higher the reconstructed image quality is. 
We find out that the model tends to have a low bit-rate proportion for the middle-frequency components when its compression ratio is high. We infer the reason is that under the strict bit-rate constriction, the learning-based model prefers to learn frequency-polarized information (i.e., giving more weights to lower and higher frequency corresponding features instead of the middle ones).

Table \ref{table:frequency-split-metrics} reveals that middle- and high-frequency components provide similar contributions to the objective metrics (i.e., PSNR and MS-SSIM) but they are different in visualization in the previous analysis as Fig.~\ref{fig:frequency_display}(b) shown.

\subsubsection{Subjective Frequency Analysis}
\label{sec:Subjective Frequency Analysis}

We could find that our low-frequency components result in a grayscale-like image, which is similar to broadly used color space Y'CbCr's luma (Y') component illustrated in Fig.~\ref{fig:08_23_display}, satisfying the feature of the human visual system (HVS).

We examine the frequency energy distribution of original images with Fourier transform and compare it with the bit-rate percentage distribution of our method. Fig.~\ref{fig:frequency_analysis}(a) shows that there exists a similar distribution that we can verify that our method based on frequency-oriented transform can learn the layered representation maintaining the same frequency energy distribution of the original images.

In addition, we evaluate each frequency component layer's contribution to the reconstruction of image quality from both subjective and objective aspects as shown in Fig.~\ref{fig:08_23_display}. 

Because of the diversity of image contents that different images' bit-rate allocation for each frequency component is different, we take all test images in the mean average calculation.
We compare each frequency split's bit-rate proportion of the whole image bit-rate percentage in Fig.~\ref{fig:frequency_analysis}(b), finding that \textit{kodim08} gives more weight to the high-frequency component while less to lower ones compared with \textit{kodim23}. 
This finding is consistent with their content difference and their frequency interpretation. Also, this contribution imbalance could be explained by the computation complexity comparison of each layer shown in Table \ref{table:frequency-layer-params} as middle- and high-frequency branches are comparable on both metrics of Macs and \#Params.

\begin{table}[t]
\centering

\caption{Evaluation the computation complexity of the encoder for each frequency split. }
\begin{tabular}{@{}llll@{}}
\toprule
Metric & Low & Middle & High \\ \midrule
Macs & 39,301M & 2,100M & 2,478M \\ \midrule
\#Params & 11.56M & 10.19M & 10.56M \\ \bottomrule
\end{tabular}

\label{table:frequency-layer-params}
\end{table}

\section{Visual Analysis}
\label{sec:visual}

To prove our proposed end-to-end optimized image compression model could effectively retain conceptual information we select two representative visual analysis tasks: object detection, and segmentation. Moreover, the proposed method supports scalable image coding by transmitting selective frequency components while acquiring comparable visual analysis performance.

We randomly select 1,000 images in the COCO2017 validation dataset \cite{lin2014microsoft} as the test dataset to evaluate object detection and segmentation tasks on it. Next generation codec H.266/VVC is chosen as the comparison method. Detailedly, we use VTM 11.0 for VVC and set all-intra mode with 8-bit YCbCr 4:4:4.  
To be noted, we adopt the padding operation on the original images before compression and cut out those paddings before conducting visual analysis tasks on them because the proposed compression model which is based on convolution operations can only handle images with sizes of the multiple of 64. 

\subsection{Object Detection}
We use an open-sourced pre-trained YOLOv5 model \cite{yolov5} for fair comparison and report the mAP@0.5.
As illustrated in Fig.~\ref{fig:multiple-tasks}(a), results show that the object detection task performs better on our model's reconstructed images compared with VVC, especially when its bit rate is less than $0.4$ bpp.
Moreover, under a high compression ratio situation, the proposed compression model shows feasibility in the incremental transmission that \{low + middle\} combination has comparable and even better performance, which illustrates the proposed model's potential for scalable encoding.

\subsection{Segmentation}

We compare the segmentation task result to the reconstructed images using the same pre-trained model of DeepLab v2 \cite{Chen2018DeepLabSI}. We adopt the dense pixel-level annotations from the COCO-Stuff \cite{Caesar2018COCOStuffTA} dataset which has 172 classes in total. Frequency weighted IoU score is reported.
It is shown in Fig.~\ref{fig:multiple-tasks}(b) that the segmentation task results better on our reconstructed images than VVC on all rates. Take the highest compression ratio point of the proposed model as an example that 2.6\% segmentation gain is achieved. Scalable coding supports the segmentation task when the bit rate is higher, which is contrary to the detection task.

In general, decoded images from the proposed model could better support both detection and segmentation tasks, which are representative tasks of computer vision.
We analyze this because our proposed method and other learning-based image compression approaches could restrain more semantic information at even high compression ratio situations, while the hand-crafted rules in traditional codecs expose their shortcomings. The scalable coding property of our model is also proved by its supportability for visual analysis tasks on different combinations of frequency splits.

\section{Conclusion}
\label{sec:conclusion}

We propose an end-to-end image compression model using the frequency-oriented transform. By examining compression degradation varies on different frequencies, we combine the idea of the spatially Laplace pyramid into image signal decomposition and design the frequency-aware fusion module. Thus our model offers interpretability from the aspect of frequency decomposition.

The comparison results, both objective and subjective, demonstrate the superior performance of the proposed method, surpassing the next-generation codec H.266/VVC in terms of MS-SSIM metric.

Moreover, we explore the interpretability of our model by quantitative measurements and intuitive visualizations. Multiple visual analysis tasks prove that the proposed model could retrain semantic-related information and scalable coding is achieved by selectively transmitting parts of the frequency components.

\section*{Acknowledgement}
The authors would like to thank the associate editor and anonymous reviewers for their constructive comments to improve the quality of this paper.
The authors thank Prof.Siwei Ma (Peking University) for valuable discussion and support. The authors thank Dr.Chuanmin Jia (Peking University) for assistance in experiment set-up and comments on the manuscript. 

\section*{Data Availability Statements}
All data generated or analysed during this study are included in this published article (and its supplementary information files).

\bibliography{refs}

\end{document}